\pdfoutput=1 

\documentclass{article} 
\usepackage{iclr2024_conference,times}

\usepackage{hyperref}
\usepackage{url}
\usepackage[utf8]{inputenc} 
\usepackage[T1]{fontenc}    
\usepackage{booktabs}       
\usepackage{amsfonts}       
\usepackage{nicefrac}       
\usepackage{microtype}      
\usepackage{xcolor}         
\usepackage{multirow}
\usepackage{amsmath}
\usepackage{graphicx}
\usepackage{colortbl}

\title{MGI: \underline{Multimodal} Contrastive pre-training of \underline{Genomic} and Medical \underline{Imaging}}

\iclrfinalcopy
\author{Jiaying Zhou \textsuperscript{*} \\
Department of Ophthalmology and Visual Sciences\\
The Chinese University of Hong Kong\\
\texttt{jiayingzhou2127@outlook.com} \\
\And
Mingzhou Jiang \textsuperscript{*}  \\
Tsinghua Shenzhen International Graduate School \\
Tsinghua University \\
\texttt{011122zmj@gmail.com} \\
\And
Junde Wu \textsuperscript{*, project lead} \\
National University of  Singapore\\
University of Oxford \\
\texttt{jundewu@ieee.org} \\
\And
Jiayuan Zhu \\
University of Oxford \\
\And
Ziyue Wang \\
National University of Singapore\\
\And
Yueming Jin \\
National University of  Singapore\\
\texttt{ymjin@nus.edu.sg} \\
}

\begin{document}

\maketitle
\begin{abstract}
Medicine is inherently a multimodal discipline. Medical images can reflect the pathological changes of cancer and tumors, while the expression of specific genes can influence their morphological characteristics. However, most deep learning models employed for these medical tasks are unimodal, making predictions using either image data or genomic data exclusively. In this paper, we propose a multimodal pre-training framework that jointly incorporates genomics and medical images for downstream tasks. To address the issues of high computational complexity and difficulty in capturing long-range dependencies in genes sequence modeling with MLP or Transformer architectures, we utilize Mamba to model these long genomic sequences. We aligns medical images and genes using a self-supervised contrastive learning approach which combines the Mamba  as a genetic encoder and the Vision Transformer (ViT) as a medical image encoder. We pre-trained on the TCGA dataset using paired gene expression data and imaging data, and fine-tuned it for downstream tumor segmentation tasks. The results show that our model outperformed a wide range of related methods.
\end{abstract}

\section{Introduction}

The medical field encompasses a variety of data modalities. Different types of medical data can reflect various disease characteristics and they are often complementary. For example, in tumor diagnosis, imaging data reveal the morphological characteristics and growth patterns of tumors, while genetic data provide information on disease susceptibility and potential biomarkers. Specific gene expressions can influence the morphological features of tumors, such as shape, size, and location. Therefore, integrating these different but complementary data can help clinicians develop a more comprehensive understanding of disease characteristics.

Most deep learning methods applied in the medical field today are unimodal, analyzing disease characteristics presented by single-modal data. Tumor formation and development are often closely related to specific gene expressions. Therefore, relying solely on imaging and ignoring molecular biology data may lead to misdiagnosis. For instance, a doctor might diagnose a tumor based only on imaging data, whereas gene expression data do not support this conclusion, resulting in a false positive—believing a tumor exists when it does not.

To address this issue, constructing methods that integrate multimodal data can learn information from different types of data, helping the model make more accurate judgments. This approach helps reduce the partial insights that come from relying on single-modal data. Recently, an increasing number of studies in the medical field have utilized deep neural networks to jointly learn from both gene and image data. For example, \citep{gundersen2020end}, \citep{venugopalan2021multimodal}, \citep{fujinami2021prediction} have attempted to use the combination of gene and image data for comprehensive diagnosis in the medical field. These multimodal paradigms typically use different encoders to separately encode images and genes. The gene encoders commonly employed are MLP or Transformer like\citep{kirchler2022transfergwas}, while the image encoders are usually CNN or ViT like \citep{yang2021multi}.

However, gene sequences are typically long sequences. MLP lacks the capability to model long sequences as it processes each input independently without considering relationships between inputs, like \citep{taleb2022contig}, thus failing to effectively capture long-range dependencies in genes sequence. The core of the Transformer is the attention mechanism, which, despite its excellent performance in capturing sequence dependencies, has a quadratic computational complexity. This results in extremely high computational and memory consumption when processing very long gene sequences, making it impractical for real-world applications, such as \citep{abdine2024prot2text}. Therefore, it is needed to develop more efficient methods to learn the long-range dependencies in long sequence data, such as gene expression.

In this paper, we propose a multimodal framwork named MGI (\underline{Multimodal} Contrastive Learning of \underline{Genomic} and Medical \underline{Imaging}), which combines gene sequence expression and medical imaging data. MGI uses Mamba to model gene sequences, leveraging its strengths in long sequence modeling to enhance the gene modeling capability within the multimodal framework. Simultaneously, it employs ViT to model images and integrates information from gene sequence expression data to achieve more accurate medical image segmentation. During the pre-training phase, we use a self-supervised contrastive learning strategy to align the visual encoder and gene encoder on paired genetic data and image data, enabling the visual encoder to learn relevant features from the perspective of genes. For the downstream segmentation task, we introduce a lightweight multimodal attention fusion decoder to integrate image and gene data, thereby improving segmentation accuracy. 

Our contributions can be summarized as follows:

(1) We propose a new multimodal model called MGI, which is based on both genetic and imaging data. We use Mamba for feature extraction from the gene data, successfully addressing the issue faced by previous gene encoders in capturing long-range dependencies in long gene sequences.

(2) We align image embeddings with gene embeddings through a gene-image contrastive loss, enabling the image encoder to understand genetic information. 

(3) We propose a new lightweight multimodal fusion module for downstream tasks. 

\section{Method}
\label{gen_inst}

Before introducing our method, we first motivate the imaging and genetic modalities chosen in this work in Sec 2.1. Then, we describe our multimodal pre-training framework MGI in Sec 2.2. After that, in Sec 2.3, we introduce the contrastive loss which aligns the images and genes. Finally, in Sec 2.4, we describe the multimodal fusion method that we design for transferring  learning of downstream segmentation task.

\subsection{Modalities of Imaging and Genetic Data}\label{2.1}

\paragraph{Medical imaging modality.} Magnetic resonance imaging (MRI), especially the fluid-attenuated inversion recovery (FLAIR) sequence, is of significant importance in the diagnosis and monitoring of lower-grade gliomas. The FLAIR sequence can suppress cerebrospinal fluid signals, thereby more clearly displaying the lesion areas and aiding physicians in identifying and evaluating the extent and location of the tumor. Therefore, in this paper, we chose the FLAIR sequence images from The Cancer Genome Atlas (TCGA) lower-grade glioma dataset as the image modality for pre-training.

\paragraph{Genetic modality.} 
The human genome consists of over 3 billion base pairs. Genes are made up of different numbers of base pairs arranged in specific sequences. Due to the limitations of sequencing technology and cost, only a small portion of genes have been sequenced, but these sequences are still quite long by current standards. Human complex traits are influenced by the expression of numerous genes; for instance, the development of Alzheimer's disease and endometrial cancer is associated with specific gene segments\citep{pietzner2021mapping}.

Therefore, when using deep learning methods to analyze medical images for disease diagnosis, it is beneficial to consider different levels of gene expression. This approach helps in understanding the impact of specific genes on the morphology of pathological regions in the images, thereby constructing more reliable and accurate models. For our pre-training, we choose the gene expression data from RNA-sequence. This data is collected from lower-grade glioma samples where mRNA is reverse-transcribed into complementary DNA, reflecting the expression levels of different gene segments during sequencing. The expression of certain genes is associated with the disease.

\subsection{Multimodal pre-training framework}\label{2.2}

\begin{figure}[h]
    \begin{center}
    \includegraphics[width=\textwidth]{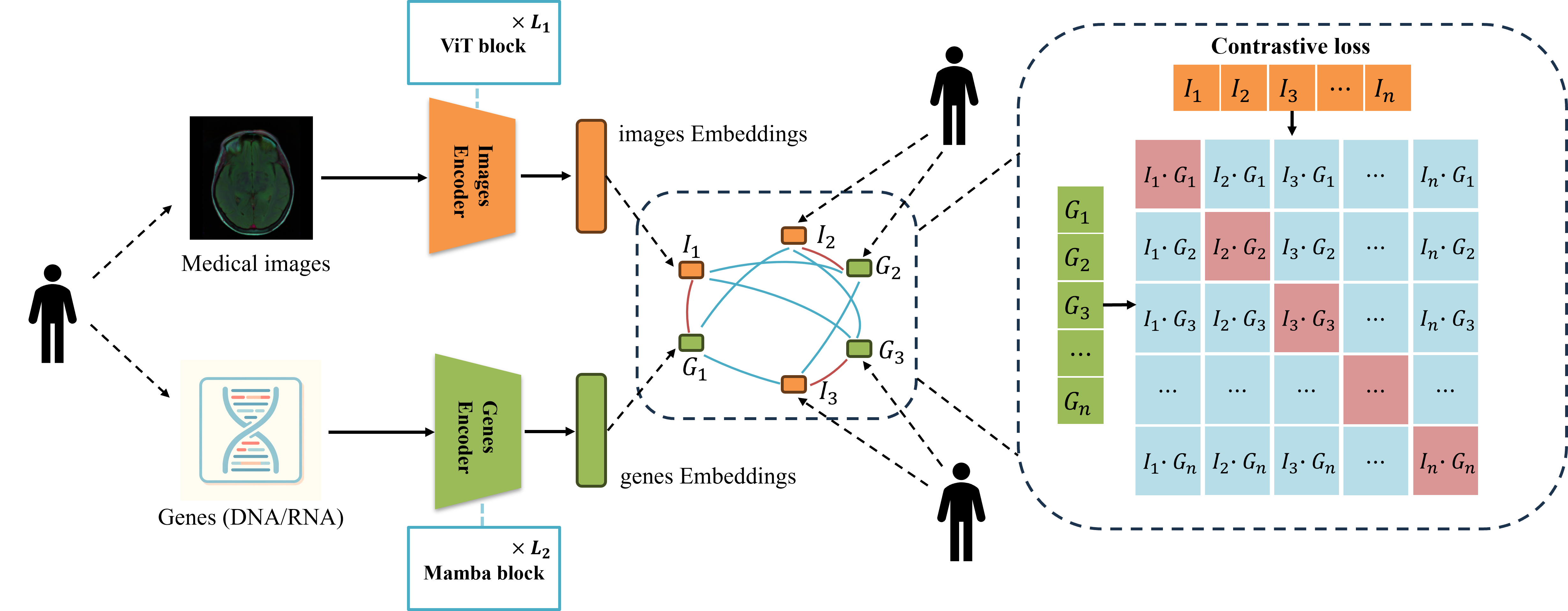}
    \end{center}
    \caption{MGI framework.}\label{fig1}
\end{figure}

As shown in Fig.~\ref{fig1}, our MGI framework consists of three parts: the inputs of multimodal samples, encoders of medical images and genes, contrastive loss for images and genetics. We assume a bacth input including \textit{B} multimodal samples, one for each individual person. There is a single medical image and the gene expression sequences corresponding to the patient in each sample. 

To process these different modalities of inputs, we use one Images Encoder and one Genes Encoder to encoding images and genes separately. We seek to exploit the Vision Transformer (ViT)'s capability to capture global image features for encoding purposes, \citep{dosovitskiy2021image} aiming to enhance the scalability and applicability of image embeddings for various downstream tasks. Simultaneously, we highly endorse Mamba's approach of utilizing state to memory the global information, which demonstrates superior long-sequence modeling capabilities and higher computational efficiency \citep{gu2023mamba}.  Therefore, in our method, the Images Encoder consists of $L_1$ Vision Transformer(ViT) blocks and Genes Encoder consists of $L_2$ Mamba blocks.

Images and genetic sequences are encoded into embeddings by their respective encoders. Then, mixed pooling and linear projection are applied to the embeddings, mapping the image and genetic embeddings into vectors within the feature space. We denote imaging feature vector by $I_i$ and genes sequences feature vector by $G_i$ ($i\in 1,2,3,..., B$). Feature vectors $I_i$  and $G_i$ from different samples are used to calculate contrastive loss, and backpropagation is employed to update the model parameters.

\subsection{Contrastive loss for Images and Genetics}\label{2.3}

We present the images-genes contrastive loss to align the images embeddings and genes embeddings. Given the imaging feature vector  $I_i$ and genes sequences feature vector $G_i$, each $I_i$ and $G_i$ is computed for cosine similarity with the all of the $G_i$ and $I_i$ within the same batch, respectively. As shown in Fig.~\ref{fig1}, we aim to maximize the cosine similarity between $I_i$ and $G_i$ from the same sample. The loss function is shown in formulation (1).
\begin{equation}
    \mathcal{L}\left ( I,G\right )= \displaystyle\sum_{i=1}^{B} log\frac{exp\left ( \cos \left ( {I}_{i},{G}_{i}\right )/ \tau \right )}{\displaystyle\sum_{j=1}^{B}\displaystyle\sum_{k=1}^{B}exp\left ( \cos \left ( {I}_{j},{G}_{k}\right )/ \tau \right )}
\end{equation}

where $\tau$ is a temperature parameter, cos is the cosine similarity. This formulation ensures that the learned visual representations and genetic sequence representations acquire useful information from each other.

\begin{figure}[h]
    \begin{center}
    \includegraphics[width=\textwidth]{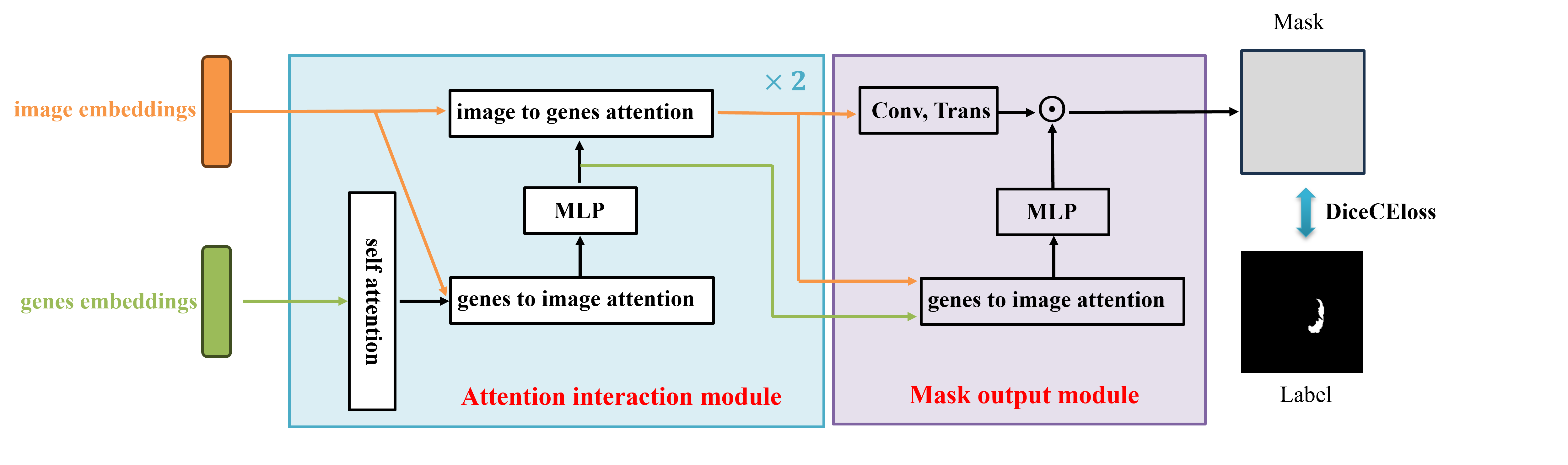}
    \end{center}
    \caption{Multimodal fusion module.}\label{fig2}
\end{figure}
\subsection{Multimodal fusion for transferring learning}\label{2.4}

There exits a gap between medical images and genetic expression sequence modalities. Our motivation for multimodal fusion is that the inter-modality interactions between medical images and genetic features would be able to enhance the effectiveness of visual representations. For example, after integrating multimodal genetic and image data, the performance of Alzheimer's disease diagnostic classification shows significant improvement\citep{zhou2019effective}. 

We present a lightweight Multimodal fusion module as the maskdecoder for downstream segmentation task. The Multimodal fusion module uses the two-way attention mechanism to integrate images embeddings and genes embeddings. It consists of two Attention interaction modules and a Mask output module. As shown in Fig.~\ref{fig2}, each Attention interaction module performs 4 steps:(1) self attention on the genes embeddings, (2) cross attention from genes embeddings to image embeddings, (3) a MLP to update features, (4) cross attention from image embeddings to genes embeddings. Finally, in Mask output module, we upsample the updated image embeddings by transposed convolutional layer, and a MlP block after one cross attention from genes embeddings to image embeddings is applied to update genes embeddings matching the dimension of image embeddings. We pre-
dict a segmentation mask with a spatially point-wise product between the image embeddings and genes embeddings.

\section{Experiment}
\label{headings}

\subsection{Dataset}

We pretrained our models on the TCGA-LGG dataset, a genomic and clinical dataset for Low-Grade Glioma collected by The Cancer Genome Atlas (TCGA) project. This dataset contains multimodal data for nearly 530 individuals. Since only a subset of the image data in LGG is available, we can only use that subset in our experiment. It includes 3,929 CT images and their corresponding masks from 106 patients. Each patient's genetic data includes the expression levels of 60,484 genes, which have been screened by doctors and are considered potentially related to low-grade glioma. We hold out a train split (80\%) from the LGG dataset, using the remaining (20\%) for testing. For the downstream task, we fine-tuned MGI for tumor segmentation task on the TCGA-LGG.

\subsection{Results}

We compare our MGI with several baseline methods in Tab.~\ref{tab1}. Our method outperforms other approaches, achieving a Dice score of 0.901.

\begin{table}[t]
\caption{The results of segmentation on the TCGA-LGG by Dice score}\label{tab1}
\centering
\begin{tabular}{cc}
\hline
 &  \\
\multirow{-2}{*}{Method} & \multirow{-2}{*}{TCGA-LGG} \\ \hline
U-Net & 0.735 \\
ResU-Net & 0.772 \\
DeepLabv3+ & 0.840 \\
Attention U-Net & 0.830 \\
\rowcolor[HTML]{ECF4FF} 
\textbf{MGI(ours)} & \textbf{0.901} \\ \hline
\end{tabular}
\end{table}

\section{Conclusion}

In these paper, we present a new multimodal framework of genomic and medical image, which utilize ViT as the image encoder and Mamba as the genes encoder. This is the first attempt at multimodal learning using ViT and Mamba. We design a image-gene contrastive loss to align image embeddings and genes embeddings. For downstream segmentation task, we present a new lightweight multimodal fusion module to integrate the features of genetic and medical image. Our method performs well on the Low-Grade Glioma segmentation task.

\bibliography{iclr2024_conference}

\begin{thebibliography}{11}
\providecommand{\natexlab}[1]{#1}
\providecommand{\url}[1]{\texttt{#1}}
\expandafter\ifx\csname urlstyle\endcsname\relax
  \providecommand{\doi}[1]{doi: #1}\else
  \providecommand{\doi}{doi: \begingroup \urlstyle{rm}\Url}\fi

\bibitem[Abdine et~al.(2024)Abdine, Chatzianastasis, Bouyioukos, and Vazirgiannis]{abdine2024prot2text}
Hadi Abdine, Michail Chatzianastasis, Costas Bouyioukos, and Michalis Vazirgiannis.
\newblock Prot2text: Multimodal protein’s function generation with gnns and transformers.
\newblock In \emph{Proceedings of the AAAI Conference on Artificial Intelligence}, volume~38, pp.\  10757--10765, 2024.

\bibitem[Dosovitskiy et~al.(2021)Dosovitskiy, Beyer, Kolesnikov, Weissenborn, Zhai, Unterthiner, Dehghani, Minderer, Heigold, Gelly, Uszkoreit, and Houlsby]{dosovitskiy2021image}
Alexey Dosovitskiy, Lucas Beyer, Alexander Kolesnikov, Dirk Weissenborn, Xiaohua Zhai, Thomas Unterthiner, Mostafa Dehghani, Matthias Minderer, Georg Heigold, Sylvain Gelly, Jakob Uszkoreit, and Neil Houlsby.
\newblock An image is worth 16x16 words: Transformers for image recognition at scale, 2021.

\bibitem[Fujinami-Yokokawa et~al.(2021)Fujinami-Yokokawa, Ninomiya, Liu, Yang, Pontikos, Yoshitake, Iwata, Sato, Hashimoto, Tsunoda, et~al.]{fujinami2021prediction}
Yu~Fujinami-Yokokawa, Hideki Ninomiya, Xiao Liu, Lizhu Yang, Nikolas Pontikos, Kazutoshi Yoshitake, Takeshi Iwata, Yasunori Sato, Takeshi Hashimoto, Kazushige Tsunoda, et~al.
\newblock Prediction of causative genes in inherited retinal disorder from fundus photography and autofluorescence imaging using deep learning techniques.
\newblock \emph{British Journal of Ophthalmology}, 105\penalty0 (9):\penalty0 1272--1279, 2021.

\bibitem[Gu \& Dao(2023)Gu and Dao]{gu2023mamba}
Albert Gu and Tri Dao.
\newblock Mamba: Linear-time sequence modeling with selective state spaces.
\newblock \emph{arXiv preprint arXiv:2312.00752}, 2023.

\bibitem[Gundersen et~al.(2020)Gundersen, Dumitrascu, Ash, and Engelhardt]{gundersen2020end}
Gregory Gundersen, Bianca Dumitrascu, Jordan~T Ash, and Barbara~E Engelhardt.
\newblock End-to-end training of deep probabilistic cca on paired biomedical observations.
\newblock In \emph{Proceedings of The 35th Uncertainty in Artificial Intelligence Conference}, 2020.

\bibitem[Kirchler et~al.(2022)Kirchler, Konigorski, Norden, Meltendorf, Kloft, Schurmann, and Lippert]{kirchler2022transfergwas}
Matthias Kirchler, Stefan Konigorski, Matthias Norden, Christian Meltendorf, Marius Kloft, Claudia Schurmann, and Christoph Lippert.
\newblock transfergwas: Gwas of images using deep transfer learning.
\newblock \emph{Bioinformatics}, 38\penalty0 (14):\penalty0 3621--3628, 2022.

\bibitem[Pietzner et~al.(2021)Pietzner, Wheeler, Carrasco-Zanini, Cortes, Koprulu, W{\"o}rheide, Oerton, Cook, Stewart, Kerrison, et~al.]{pietzner2021mapping}
Maik Pietzner, Eleanor Wheeler, Julia Carrasco-Zanini, Adrian Cortes, Mine Koprulu, Maria~A W{\"o}rheide, Erin Oerton, James Cook, Isobel~D Stewart, Nicola~D Kerrison, et~al.
\newblock Mapping the proteo-genomic convergence of human diseases.
\newblock \emph{Science}, 374\penalty0 (6569):\penalty0 eabj1541, 2021.

\bibitem[Taleb et~al.(2022)Taleb, Kirchler, Monti, and Lippert]{taleb2022contig}
Aiham Taleb, Matthias Kirchler, Remo Monti, and Christoph Lippert.
\newblock Contig: Self-supervised multimodal contrastive learning for medical imaging with genetics.
\newblock In \emph{Proceedings of the IEEE/CVF Conference on Computer Vision and Pattern Recognition}, pp.\  20908--20921, 2022.

\bibitem[Venugopalan et~al.(2021)Venugopalan, Tong, Hassanzadeh, and Wang]{venugopalan2021multimodal}
Janani Venugopalan, Li~Tong, Hamid~Reza Hassanzadeh, and May~D Wang.
\newblock Multimodal deep learning models for early detection of alzheimer’s disease stage.
\newblock \emph{Scientific reports}, 11\penalty0 (1):\penalty0 3254, 2021.

\bibitem[Yang et~al.(2021)Yang, Belyaeva, Venkatachalapathy, Damodaran, Katcoff, Radhakrishnan, Shivashankar, and Uhler]{yang2021multi}
Karren~Dai Yang, Anastasiya Belyaeva, Saradha Venkatachalapathy, Karthik Damodaran, Abigail Katcoff, Adityanarayanan Radhakrishnan, GV~Shivashankar, and Caroline Uhler.
\newblock Multi-domain translation between single-cell imaging and sequencing data using autoencoders.
\newblock \emph{Nature communications}, 12\penalty0 (1):\penalty0 31, 2021.

\bibitem[Zhou et~al.(2019)Zhou, Thung, Zhu, and Shen]{zhou2019effective}
Tao Zhou, Kim-Han Thung, Xiaofeng Zhu, and Dinggang Shen.
\newblock Effective feature learning and fusion of multimodality data using stage-wise deep neural network for dementia diagnosis.
\newblock \emph{Human brain mapping}, 40\penalty0 (3):\penalty0 1001--1016, 2019.

\end{thebibliography}
\bibliographystyle{iclr2024_conference}


\end{document}